# Efficient logic architecture in training gradient boosting decision tree for high-performance and edge computing

**Takuya Tanaka, Ryosuke Kasahara, Daishiro Kobayashi**
Ricoh Institute of Information and Communication Technology,
Research and Development Division, Ricoh Company,
2-7-1 Izumi, Ebina, Kanagawa 243-0460, Japan
takuya.tt.tanaka@jp.ricoh.com

## Abstract

This study proposes a logic architecture for the high-speed and power efficiently training of a gradient boosting decision tree model of binary classification. We implemented the proposed logic architecture on an FPGA and compared training time and power efficiency with three general GBDT software libraries using CPU and GPU. The training speed of the logic architecture on the FPGA was 26-259 times faster than the software libraries. The power efficiency of the logic architecture was 90-1,104 times higher than the software libraries. The results show that the logic architecture suits for high-performance and edge computing.

## 1     Introduction

Machine learning has recently undergone rapid development and is widely used. In general, the machine learning training processes for large data take a very long time; therefore, high-speed training is desired. In addition, the prediction performance may decrease when the prediction target changes over time [1], and training time may be an issue when tracking changes over a short period of time. Further, for training in edge devices, which have a power supply limitation, the power efficiency of training is important. Gradient boosting decision tree (GBDT), a high-performance algorithm [2], has various libraries [3]–[5] that are widely used for data analysis competitions [3], [6], [7]. Training time is also an important factor when using GBDT. For the reduction of training time, various methods, such as compress feature values, reducing calculation volume [4], and employing GPU [8]–[10] were proposed.

Apart from GBDT, the convolutional neural network (CNN) [11] is widely known as a high-performance machine learning algorithm [11]. CNN performs convoluted calculations in parallel; therefore, the GPU performs training extremely fast. For example, AlexNet, a typical CNN model with four hidden layers, was 23–206 times faster [12] compared with cases wherein 32 threads of a Xeon CPU E5-2630 v3 2.40 GHz, Intel, and GTX 1080, Nvidia, were used for training. In general, the training of GBDT is faster than that of CNN for the same task. However, regarding the reduction of training time using GPU, GBDT is not successful compared with CNN. Compared with 40 threads of a Xeon E5-2640 v4 2.4 GHz, Intel, and a high-end Tesla P100, Nvidia, GBDT was only up to 1.87 times faster [8].

GBDT is a machine learning model built from multiple decision trees. Due to the nature of the algorithm, the decision trees often have random access to training data. When using a CPU or GPU, training data is allocated to dynamic random access memory (DRAM). However, DRAM has a high random access latency, which becomes a bottleneck when increasing the training speed. Static random access memory (SRAM) has low random access latency. The development of logic architecture for GBDT training, which stores training data in SRAM and utilizes the low latency of SRAM, holds promise for increasing the training speed.

Random forest, which comprises a large number of decision trees using FPGA, has high training speed [13]. Compared with training using the Core i5 3.2 GHz, Intel, random forest has increased speeds by up to 211 times. In contrast to random forest, which learns each decision tree in parallel, GBDT learns each decision tree additively, one at a time. Therefore, GBDT training speed cannot be increased using the configuration proposed in a previous study [13].

High-speed training with the same power consumption also improves power efficiency per sample. Therefore, if the training speed of a logic architecture for GBDT is significantly faster than CPU and GPU, the power consumption of the logic architecture is superior to that of the CPU and GPU. This study proposes a novel logic architecture for GBDT training and compares the performances of the proposed logic architecture with that of the CPU and GPU of multipurpose devices.

## 2    Logic architecture for training of gradient boosting decision tree

The RAM bandwidth of an FPGA when using Virtex UltraScale+ VU9P, Xilinx is ~800 GB/s at a clock frequency of 100 MHz. In addition, the RAM on FPGA can operate at above 100 MHz. In contrast, in the case of RAM connected to a CPU, the bandwidth of one DIMM is limited to 25.6 GB/s with the current generation DDR4. Also, the GDDR5X memory is used for GPU results in a greater bandwidth compared with when using DDR4. For example, with NVIDIA GEFORCE GTX 1080 Ti, the bandwidth is 484 GB/s. Thus, The RAM on the FPGA has better bandwidth than the external memory on CPU and GPU.

Thus far, we have only discussed sequential access to addresses; however, random access has a much larger access time compared with that of sequential access in DRAM. Since the RAM on the FPGA are SRAM, the access latency is 1 clock for sequential and random access. DDR4 and GDDR5X is DRAM, and the latency increases for random access. Therefore, for GBDT, which requires decision trees with a large degree of random access, SRAM increases the training speed more effectively. Based on the precondition that the training data is placed in RAM on FPGA, FPGA is more effective than CPU and GPU at speeding up training of GBDT.

Herein, we describe the training engine, which is the basic architecture for GBDT training, and the data parallel that assigns training data to multiple training engines and partitions to speed up training. GBDT comprises functions such as binary classification, multiclass classification, regression, and ranking; only binary classification is considered herein. The loss function covered is the cross-entropy loss function, which is supported in GBDT software libraries. Training is processed using the exact greedy algorithm in xgboost [3], and the bit-width for the feature value is 8 bits.

### 2.1    Preprocessing of input dataset

For preprocessing, the feature value is converted to 8 bits. This processing is conducted externally to the FPGA in advance. For each training data feature value, 255 quantiles are calculated, and this is the bin centroid. For feature values with 255 patterns or less, each value is the bin centroid. One of the bins is assigned to missing values; therefore, in the case of feature values for continuous values, there is a maximum of 255 patterns. Using the bins of each feature value, each feature value for the training and validation data is allocated to the closest bin.

The preprocessing of feature values is not limited to the aforementioned method; any method that can convert a feature value to 8 bits can be used. In addition, if the feature value is converted to 8 bits, the categorical feature value can be used in the same way as continuous feature values.

## 2.2 Training engine

### 2.2.1 Overview of training engine

The training engine, which is the basic architecture that enables GBDT training, is discussed herein. Figure 1 illustrates the configuration of the training engine.

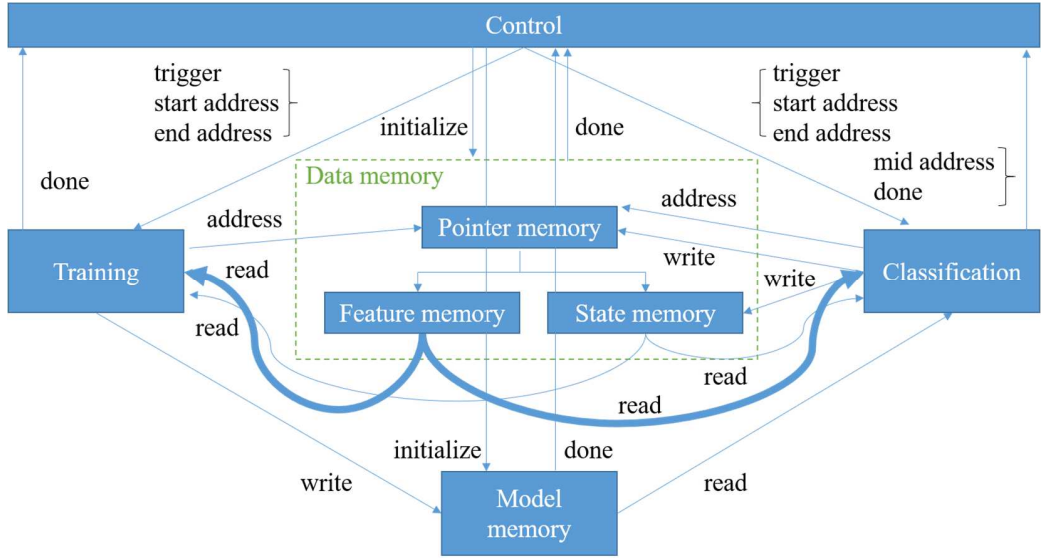

Figure 1. Overview of training engine

The training engine comprises the following five modules. The control module controls each module; the data memory module stores each type of data; the feature memory stores the feature values; and the state memory stores the sample weight, gradient information, and label data for each sample. The pointer memory stores the address table for finding all indirect addresses in the feature memory and the state memory. The model memory module stores the latest learned decision tree model. The training module trains a decision tree model for each node. The classification module determines the leaf or node for each sample data based on the trained model.

### 2.2.2 Dataflow

The flow of data at each step of training is shown as follows.

**Step 1: Start training**

When the training start signal is input to the control module, the control module sends an initialization signal (initialize in Figure 1) to the data memory module and the model memory module.

**Step 2: Initialization**

The data memory module sends an initialization signal to the pointer memory. When the initialization is completed, a done signal (done in Figure 1) is sent to the control module. The model memory module initializes RAM at each depth and sends a done signal to the control module.

**Step 3: Node training**

The start and end addresses for the pointer memory expressing the sample to learn on the current node and a trigger signal are sent from the control module to the training module. From the start address to the end address, the training module receives, via pointer memory, the feature values from feature memory, the gradient information from state memory, and the sample weight and label information for each sample. The received gradient information is sent to the gradient histogram memory prepared for each feature and added to the corresponding feature value bin.

Using the gradient information stored in the gradient histogram memory, it calculates the split gain, which expresses the value of branches for all conditions (feature and thresholds). It then calculates the feature and the threshold for which the split gain is the highest. It then calculates the flag indicating whether the node is a leaf, the branch direction in the case of a missing value. If the node is a leaf, it calculates the leaf weight. The calculated value is then sent to the model memory module, which sends the done signal to the control module.

**Step 4: Data split**

The control module sends the start and end addresses of the pointer memory, and trigger to the classification module. The classification module receives the current node branch conditions (feature, thresholds, and a flag indicating whether a node is a leaf) from the model memory module. Next, from the start address to the end address, the feature values are received from the feature memory via the pointer memory. The received feature values and branch conditions are compared and a decision is made as to which of the two lower tier nodes to a branch. The result is then sent to the pointer memory. The mid address for determining the start address and end addresses, which are required for training the lower tier node, is determined and sent to the control module.

Steps 3 and 4 are repeated for each node for the one decision tree.

**Step 5: Update gradient information**

When the training for all nodes is complete, it uses the learned decision tree model to update the sample weight and gradient information for each sample. The control module sends the start and end addresses of the feature memory and the state memory, and trigger to the classification module.

Then, the feature memory is accessed from the start address to the end address and the feature values are received. Similarly, the sample weight, gradient information, and label data are received for each sample from state memory, and leaf weight from model memory. The received leaf weight is added to the sample weight for each sample. The gradient information is calculated using the sample weight for each updated sample, and the results are sent to state memory. When processing is complete for all samples, the done signal is sent to the control module. It returns to Step 2 until the maximum number of trees is reached.

**Step 6: Training done**

The operation stops when training for the maximum number of trees is complete. When the control module receives the training start signal, it returns to Step 1.

## 2.3 Data parallel

With the data parallel, multiple training engines are used and a training sample is allocated to each training engine to speed up the training process. A configuration with two training engines is shown in Figure 2. Note that the number of engines is not limited to two, and if more FPGA resources are available, a larger number of engines can be used. Further, by tuning the number of engines, a tradeoff of the logic size, power consumption and training speed can be controlled.

The data flow for data parallel is the same, excluding Step 3 in the previous subsection. Step 3 in this configuration has the following flow.

The gradient histogram is calculated using the histogram calculation module and performed independently for each engine. The split gain calculation module adds gradient histograms for each engine and performs the split gain calculation. Then, it calculates the feature values and thresholds that make the maximum for split gain. It also calculates the flag expressing whether the node is a leaf, the branch direction in the case of a missing value, and the leaf weight in case the node is a leaf. Each value is then sent to the model memory module of each engine.

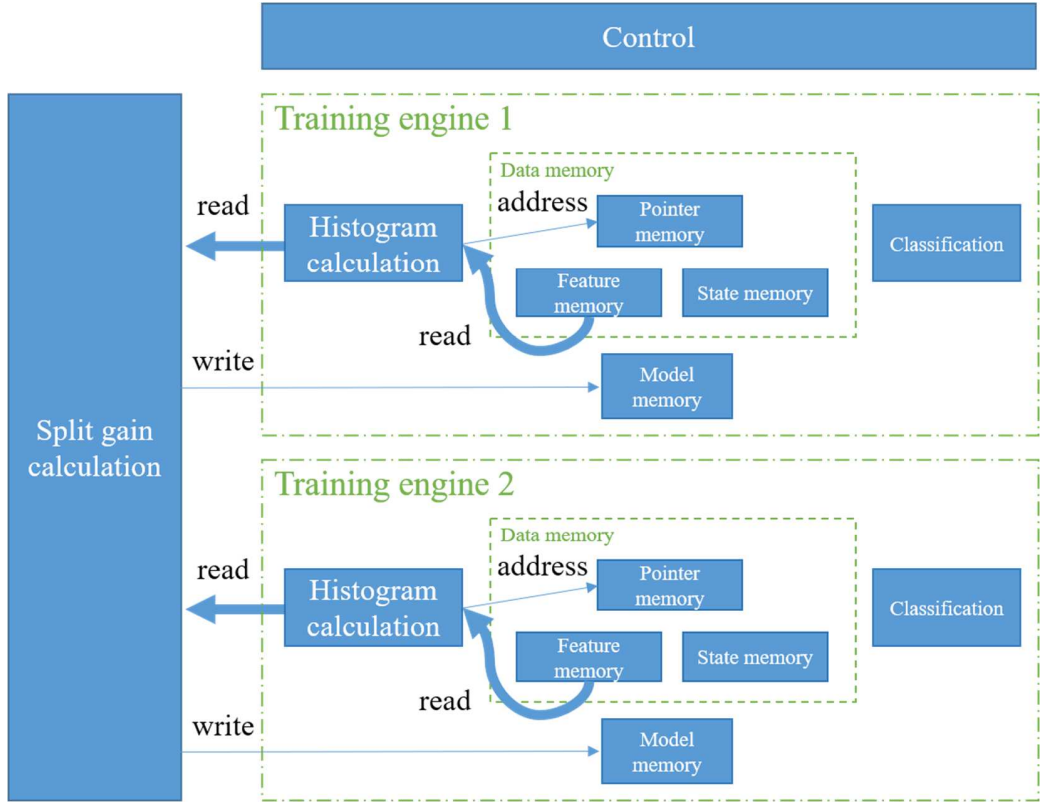

Figure 2. Overview of a data parallel model (two engines)

## 3 Method

The proposed logic architecture is implemented on the FPGA. The accuracy, training time, and power efficiency of the logic architecture are compared to the software libraries. The Higgs dataset [14] is used for the comparison.

### 3.1 Software library

As a benchmark, we used a PC with Core i7-8700K (3.70 GHz, 6 cores, 12 threads), Intel, GTX 1080Ti, Nvidia, and 64GB RAM. The three general GBDT libraries shown in Table 1 were used for the comparison. The training and validation data were converted to 8 bits in advance using the method in Section 2.1. The processing time was calculated from the time preceding and following the execution of the fit function proposed in each library.

Table 1. GBDT library

| Library | Version |
| --- | --- |
| xgboost [3] | 0.80 <pip> |
| lightgbm [4] | 2.2.0 <pip> |
| catboost [5] | 0.10.1 <pip> |

### 3.2 FPGA

An FPGA, XCVU9P-L2FLGA2104, Xilinx was used for the implementation of the logic architecture. A Virtex UltraScale+ FPGA VCU118 evaluation kit, Xilinx, was used for the FPGA board. Table 2 shows the implementation settings of FPGA. The training parameters were set to

max_depth = 1, subsample = 0.5, lambda = 1, gamma = 0, and the number of decision trees was 100.

Table 2. Implementation settings

| Number of train samples | Number of validation samples | Number of engines | Target frequency [MHz] |
|---|---|---|---|
| 10,048 | 10,048 | 64 | 100 |

### 3.3 Power consumption

For the general three libraries, using one CPU (12 threads) and GPU (1 device), we measured the FPGA's power consumption. For both the PC and FPGA, we measured the power consumption at the outlet of the PC and FPGA boards. For the measurement of each software library, we set the number of trees to 100,000 in order for the training time to continue. For the FPGA measurement, the number of trees was set to 100, and the FPGA was measured in such a way that when training stopped, retraining immediately followed. The measurement of power consumption was performed using a TAP-TST7, Sanwa Supply.

## 4 Results and Discussion

### 4.1 Implementation result

The resource usage is shown in Table 3. The clock frequency of logic is 100 MHz. There is the possibility that the clock frequency can be improved by optimizing the processing pipeline.

Table 3. Resource utilization

| Resource | Used | Available | Percentage [%] |
|---|---|---|---|
| LUT | 934,842 | 1,182,240 | 79.07 |
| LUTRAM | 343,274 | 591,840 | 58.00 |
| FF | 866,740 | 2,364,480 | 36.66 |
| BRAM | 1,829 | 2,160 | 84.68 |
| URAM | 320 | 960 | 33.33 |
| DSP | 656 | 6,840 | 9.54 |

With the FPGA, the limit on the usable samples in the Higgs dataset is ~1.08 M based on the capacity of SRAM (UltraRAM). For a larger number of samples, an FPGA with a high SRAM capacity is needed. Additionally, there are also methods involving the use of multiple FPGAs or ASIC.

### 4.2 Accuracy

We calculated the AUC for each decision tree from 1 to 100 and used a maximum value of AUC. The comparison of AUC was shown in Table 5. We confirmed that the AUC in the FPGA and in the software libraries is the same level. The results indicated that proposed logic architecture can train a GBDT model successfully.

Table 5. Comparison of AUC

|  | CPU | GPU |
|---|---|---|
| xgboost [3] | 0.7510 | 0.7518 |
| lightgbm [4] | 0.7598 | 0.7598 |
| catboost [5] | 0.7601 | 0.7592 |
| FPGA | 0.7562 ||

### 4.3 Training time

Table 6 shows the training time for three GBDT libraries and the FPGA. The numbers in parentheses in the table indicate the magnification compared with FPGA. We used the hist and gpu_hist methods for xgboost, the gbdt method for lightgbm, and the plain method for catboost. The processing time of FPGA was 26–259 times faster than that of each of the libraries under the same conditions.

Table 6. Comparison of processing times

|  | CPU [ms] | GPU [ms] |
|---|---|---|
| xgboost [3] | 107.5 ( ×44 ) | 110.2 ( ×45 ) |
| lightgbm [4] | 63.9 ( ×26 ) | 286.1 ( ×116 ) |
| catboost [5] | 432.9 ( ×176 ) | 636.2 ( ×259 ) |
| FPGA | 2.5 ||

Regarding CPU and GPU, although, they have a small amount of SRAM as a cache. However, the result also shows the fact that in decision tree training, there is no locality in access, and it is difficult to use the cache effectively.

Some of the benefits of improving training speed are as follow. The GBDT has a wide variety of training parameters. There are methods, such as Bayesian optimization, to efficiently select parameters. However, the optimization process is very time-consuming because the training of GBDT often takes a long time, and the number of training cycles is huge. Using our proposed logic architecture, the scope of parameters that can be simultaneously searched for increases, and a more accurate model can be developed. In addition, when it is necessary to update the model when the target changes with time, our proposed logic architecture can train models more often. As a result, the trained model can follow the target's change immediately, and the accuracy is improved.

### 4.4 Power consumption

Table 7 shows the power consumption for three GBDT libraries and FPGA. The power consumption per sample is shown in Table 8. The numbers in parentheses in the table indicate magnification of power consumption efficiency per a sample compared with FPGA. The power consumption per sample was calculated from the results in Tables 6 and 7. The number of samples is 10,048 for training, 10,048 for validation, and the number of decision trees is 100. Of the three GBDT libraries, the one with lightgbm (CPU) had the lowest power consumption per sample. The power consumption efficiency per sample in FPGA was 90–1,104 times higher than the GBDT libraries. The results show that the logic architecture is suitable for edge devices and computing, which require low power consumption.

Table 7. Comparison of power consumption

|  | Processing state [W] | |
|---|---|---|
|  | CPU | GPU |
| xgboost[3] | 157 | 161 |
| lightgbm[4] | 155 | 156 |
| catboost[5] | 124 | 192 |
| FPGA | 45 | |

Table 8. Comparison of power consumption efficiency per a sample

|  | CPU [ mW / sample ] | GPU [ mW / sample ] |
|---|---|---|
| xgboost[3] | 0.840 ( ×153 ) | 0.883 ( ×160 ) |
| lightgbm[4] | 0.493 ( ×90 ) | 2.22 ( ×403 ) |
| catboost[5] | 2.67 ( ×485 ) | 6.08 ( ×1,104) |
| FPGA | 0.0055 | |

## 5  Conclusion

We proposed a logic architecture to enable high-speed GBDT training. The proposed logic architecture was implemented on an FPGA, and the training time was compared with typical GBDT software library training times, confirming 26–259 times faster processing times and 90–1,104 times more efficient power consumption. Further, we confirmed that the AUC on the FPGA and in the software libraries is at the same level. These results show that the logic architecture is suitable for high-performance and edge computing. Our future research aims to extend the functionality of the GBDT, including multi-class classification, regression, and ranking, and use multiple FPGAs to increase the speed and volume of the data handled.

**Acknowledgments**

We would like to thank Takashi Totsuka, Tamon Sadasue and Akira Kinoshita for their valuable feedback.